\documentclass[journal,twoside,web]{ieeecolor}
\usepackage{cite}
\usepackage{amsmath,amssymb,amsfonts}
\usepackage{algorithmic}
\usepackage{graphicx}
\usepackage{algorithm,algorithmic}
\usepackage{hyperref}
\hypersetup{
    colorlinks=true,
    linkcolor=blue,
    citecolor=blue,
    urlcolor=blue
}
\usepackage{textcomp}
\usepackage{booktabs}
\usepackage{multirow}
\usepackage{adjustbox}
\usepackage{wrapfig}
\usepackage{placeins}
\usepackage{tikz}
\usepackage{pgfplots}
\pgfplotsset{compat=1.18}
\definecolor{academicblue}{HTML}{1F77B4}
\definecolor{academicred}{HTML}{D62728}
\def\logoname{logo}
\definecolor{subsectioncolor}{rgb}{0,0,0}
\newcommand{\best}[1]{\textbf{#1}}
\newcommand{\second}[1]{\underline{#1}}

\def\BibTeX{{\rm B\kern-.05em{\sc i\kern-.025em b}\kern-.08em
    T\kern-.1667em\lower.7ex\hbox{E}\kern-.125emX}}

\makeatletter
\def\ps@titlepagestyle{%
  \def\@oddfoot{}%
  \def\@evenfoot{}%
  \def\@oddhead{}%
  \def\@evenhead{}%
}
\makeatother

\begin{document}

\title{MOSAIC: Modality-Specific Adaptation for Incremental Continual Learning in Parkinson’s Disease Gait Assessment}

\author{Minlin~Zeng, Zhipeng~Zhou, Yang~Qiu,
Martin~J.~McKeown, and Zhiqi~Shen
\thanks{Minlin Zeng, Zhipeng Zhou, Yang Qiu, and Zhiqi Shen are with Nanyang Technological University, Singapore (e-mail: minlin001@e.ntu.edu.sg; zzpustcml@gmail.com; qiuyang@ntu.edu.sg; zqshen@ntu.edu.sg).}%
\thanks{Martin J. McKeown is with the Pacific Parkinson's Research Centre, University of British Columbia, Canada (e-mail: martin.mckeown@ubc.ca).}}

\maketitle

\begin{abstract}
Gait-based Parkinson's disease assessment increasingly relies on heterogeneous sensors, but clinical systems rarely collect all modalities simultaneously. New sensors may arrive through device upgrades, protocol changes, or multi-center deployment, while historical patient data are often unavailable because of privacy and storage constraints. This modality-incremental setting faces three challenges: unreliable cross-modal distillation, modality-specific statistical shifts, and reduced plasticity after preservation. We propose MOSAIC, a compact continual learning framework. First, we identify the \textit{Toxic Teacher} phenomenon and introduce Modality-Specific Warm-Up to stabilize newly learned modality representations before distillation. Second, we propose a statistics-decoupled MSBN architecture that isolates sensor statistics while maintaining a shared semantic backbone. Third, we design a curriculum-guided repulsive objective for Plasticity Recovery, preserving legacy knowledge while recovering modality-specific capacity. Experiments on three multimodal Parkinson's gait datasets show that MOSAIC improves final performance and mitigates forgetting. Project code is available at \href{https://github.com/minlinzeng/MOSAIC_Modality-Specific-Adaptation-for-Incremental-Continual-Learning-in-PD-Gait-Assessment.git}{GitHub}.

\end{abstract}

\begin{IEEEkeywords}
Gait analysis, Multimodal learning, Parkinson's disease.
\end{IEEEkeywords}

\section{Introduction}
\label{sec:introduction}
Parkinson's disease (PD) is a rapidly growing neurodegenerative disorder, with the global affected population projected to reach 9 million by 2030 \cite{bhidayasiri2023tackling}. The escalating shortage of specialized healthcare providers has accelerated the development of Artificial Intelligence (AI) for automated clinical assessment \cite{dorsey2018parkinson}. Because PD manifests through complex and highly variable motor dysfunctions, accurate assessment cannot rely on a single metric. Instead, multimodal gait analysis—integrating heterogeneous sensors such as depth sensor, clinical walkways, and inertial measurement units (IMUs)—has emerged as the gold standard, as different modalities capture complementary kinematic and kinetic facets of the disease \cite{zeng2025towards, rao2025survey}.

\begin{figure}[!t]
\centering
\includegraphics[width=0.98\linewidth]{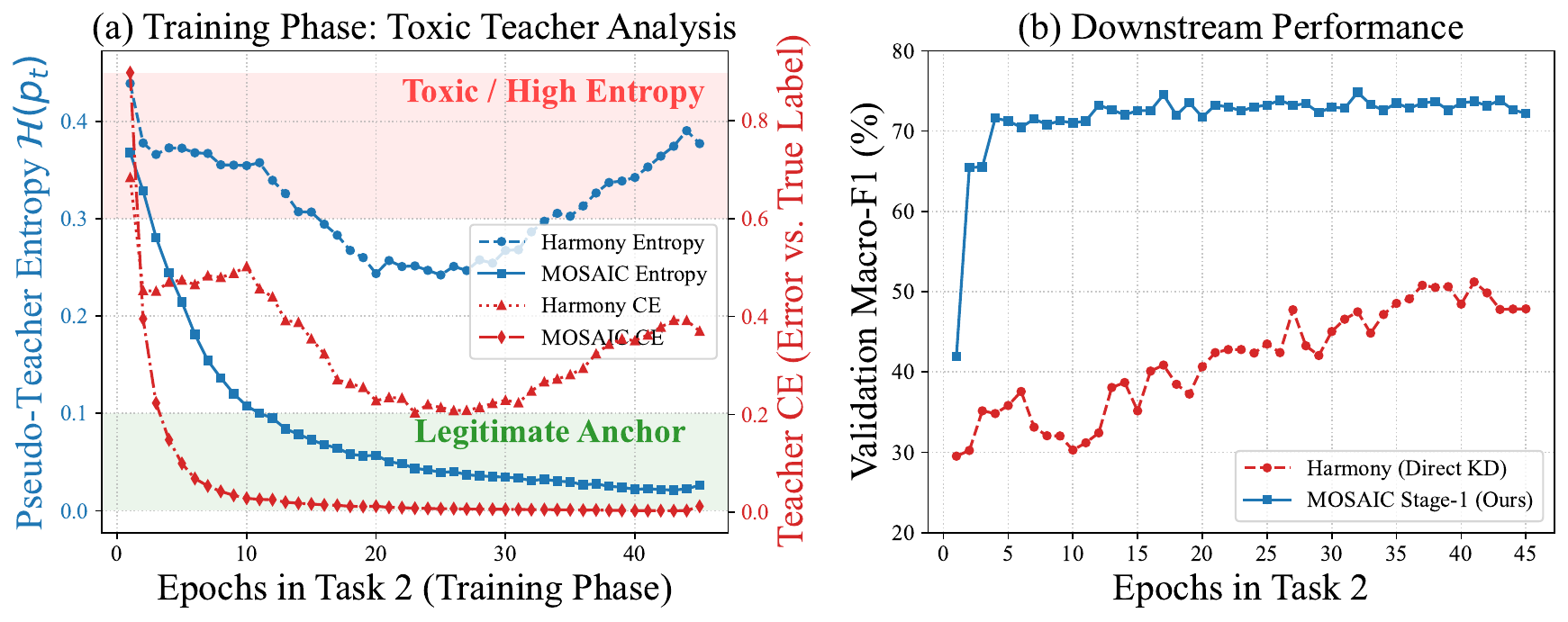}
\caption{Illustration of the "Teacher Misalignment" paradox in direct cross-modal KD. (a) Direct KD yields a pseudo-teacher with toxic, high-entropy outputs, whereas our Stage~1 Modality-Specific Warm-Up rapidly stabilizes the entropy into a legitimate anchor. (b) Forcing alignment with a high-entropy teacher causes catastrophic negative transfer, which our method successfully prevents.}
\label{fig:toxic_teacher}
\vspace{-0.8em}
\end{figure}

Despite the diagnostic advantages of multimodal learning, deploying such systems in real-world clinical artificial intelligence of things (AIoT) environments presents a critical data logistics challenge. In practice, diverse sensing modalities are rarely acquired simultaneously. New types of sensors are often introduced sequentially due to clinical upgrades, evolving diagnostic protocols, or multi-center collaborations \cite{song2025harmony}. Furthermore, strict patient privacy regulations prohibit the indefinite retention of historical physiological data \cite{pang2025breaking}, making it impossible to retrain multimodal models from scratch when a new sensor is adopted. To address this, Modality-Incremental Continual Learning (MICL) has become essential. MICL provides a framework to continually integrate novel sensor streams into a compact, unified clinical model, preserving legacy diagnostic capabilities without violating privacy constraints or incurring prohibitive retraining costs.

\begin{table*}[!htbp]
\centering
\caption{Comparison of representative class- and modality-incremental learning paradigms under criteria relevant to clinical deployment. Raw-data-free denotes that previous raw training data are not required. Pretraining-free denotes no reliance on large-scale pretrained backbones. Modality scalablity denotes that model capacity does not grow linearly with the number of modalities. Explicit statistical decoupling denotes direct separation of modality-specific normalization statistics. N/A* denotes methods that bypass shared statistics through model isolation rather than explicit statistical decoupling.}
\label{tab:related_comparison}
\scriptsize
\begin{adjustbox}{width=\textwidth}
\begin{tabular}{llp{2cm}ccccp{3.9cm}}
\toprule
\textbf{Paradigm} & \textbf{Method} & \textbf{Target Modalities} & \textbf{Raw-Data} & \textbf{Modality} & \textbf{Pretraining} & \textbf{Explicit Stat.} & \textbf{Core Mechanism} \\
 &  &  & \textbf{Free} & \textbf{Scalability} & \textbf{Free} & \textbf{Decoupling} &  \\
\midrule
Rehearsal & MedCoSS \cite{ye2024continual} & CT, MRI, X-Ray & $\times$ & $\checkmark$ & $\checkmark$ & $\times$ & Replay + KD \\
\midrule
Isolation & CMR-MFN \cite{wang2023confusion} & RGB, IMU & $\checkmark$ & $\times$ & $\checkmark$ & N/A* & Frozen modality-specific backbones \\
 & DRMN \cite{hegde2025modality} & RGB, IR, Gray & $\checkmark$ & $\times$ & $\checkmark$ & N/A* & Disjoint parameter allocation \\
 & LwI \cite{chenlearning} & RGB-only & $\checkmark$ & $\checkmark$ & $\checkmark$ & $\times$ & Graph matching + Pathway protection \\
\midrule
Foundation & CR-LoRA \cite{zhang2025contrastive} & PET, MRI, CT & $\checkmark$ & $\checkmark$ & $\times$ & $\times$ & Massive pretrained backbone + LoRA \\
 & Retinal VLP \cite{yao2025continual} & RGB, Text & $\checkmark$ & $\checkmark$ & $\times$ & $\times$ & Foundation model + Text anchors \\
\midrule
Projection & HarMI \cite{zhang2021harmi} &Acc., Gyro., Mag. & $\checkmark$ & $\checkmark$ & $\checkmark$ & $\times$ & CCA + Attention + Shared projection \\
 & Harmony \cite{song2025harmony} & RGB, Depth, IR & $\checkmark$ & $\checkmark$ & $\checkmark$ & $\times$ & Contrastive distillation, Attention \\
\midrule
Ours & MOSAIC & GRF, IMU, Skel. & \textbf{$\checkmark$} & \textbf{$\checkmark$} & \textbf{$\checkmark$} & \textbf{$\checkmark$} & MSBN + Repulsive loss + KD \\
\bottomrule
\end{tabular}
\end{adjustbox}
\end{table*}

As summarized in Table~\ref{tab:related_comparison}, although existing MICL strategies rely on memory rehearsal, parameter isolation, or foundation-model adaptation, they are fundamentally incompatible with clinical gait assessment due to strict privacy laws, lack of cross-modal knowledge transfer, and the absence of massive pre-training data for heterogeneous sensors. Consequently, latent projection without explicit structural isolation has emerged as the most viable path forward. The current state-of-the-art (SOTA) in this trajectory, Harmony \cite{song2025harmony}, pioneers cross-modal Knowledge Distillation (KD) for vision-centric modality-incremental learning (e.g., sequential integration of RGB, Depth, and Infrared (IR) images). While this approach successfully projects structurally homologous inputs into a shared semantic manifold, its core architectural assumption—that novel inputs can be seamlessly and instantly mapped into a unified abstract space—fundamentally breaks down when applied to clinical IoT applications. Unlike visual modalities that share dense 2D spatial topologies and similar statistical distributions, clinical sensors span strictly disjoint physical domains with asymmetric sampling structures and vastly different physical variances. Under such severe cross-modal heterogeneity, the direct projection mechanism advocated by recent SOTA triggers a catastrophic optimization failure we identify as the \textit{Toxic Teacher} phenomenon. Because a newly introduced, uncalibrated encoder cannot instantly bridge this extreme physical gap, it initially produces chaotic representations. As empirically demonstrated in Fig.~\ref{fig:toxic_teacher}, enforcing immediate cross-modal KD aligns the shared network with this untrained pseudo-teacher, generating a high-entropy target distribution ($\mathcal{H}(p_t) \to \max$) coupled with massive cross-entropy error. Distilling from this toxic target injects destructive gradient noise, actively degrading legacy decision boundaries rather than preserving them.

Furthermore, forcing these disjoint sensors through shared normalization layers causes catastrophic variance coupling, where historical statistical moments ($\mu, \sigma^2$) are aggressively overwritten. Finally, even if legacy topology is preserved, purely preservative distillation over-constrains the latent space, suffocating the plasticity required to capture the incoming sensor's unique physical variations. To overcome these challenges, we propose MOSAIC, a novel framework designed for modality-incremental clinical assessment. Our primary contributions are summarized as follows:
\begin{itemize}
    \item \textbf{Modality-Specific Warm-Up:} We identify the \textit{Toxic Teacher} phenomenon in cross-modal KD and stabilize newly introduced modality representations before distillation, yielding reliable pseudo-teachers for subsequent Semantic Preservation.
    \item \textbf{Statistics-Decoupled MSBN Architecture:} We introduce MSBN to isolate modality-specific statistics while preserving shared convolutional filters and a unified semantic backbone.
    \item \textbf{Plasticity Recovery:} We design a stabilize-then-expand repulsive curriculum that prevents post-distillation collapse and restores capacity for incoming modality-specific structures.
    \item \textbf{Extensive Empirical Validation:} Experiments on three multimodal PD gait datasets show that MOSAIC consistently improves final performance and reduces forgetting across modality arrival orders.
\end{itemize}

\section{Related Work}
\label{related_work}
\subsection{Multimodal Gait-based PD Assessment}
In gait-based PD assessment, multimodal analysis has become an important paradigm for capturing complementary disease manifestations from heterogeneous sensory data \cite{zhao2025artificial}. Existing approaches commonly model cross-modal dependencies through early or late fusion \cite{zhao2021multimodal}, graph-based shared latent representations \cite{hu2021graph}, and cross-attention mechanisms \cite{cui2023multi}. These methods improve representation learning when all target modalities are available during training, but this assumption is restrictive in realistic clinical settings where sensor availability, data-sharing permissions, and acquisition protocols may evolve over time. Even methods designed to tolerate flexible or missing inputs at inference time, such as TRIP \cite{zeng2025towards}, still require concurrent multimodal observations and multitask learning paradigm \cite{zhou2026exploring, zhou2025injecting} during training to construct joint representations. Therefore, conventional multimodal PD assessment methods are not directly suited to scenarios in which new sensor types must be incorporated sequentially without access to historical paired data.

\subsection{Modality Incremental Continual Learning}
The objective of MICL is to sequentially integrate heterogeneous modalities into a unified model while retaining previously learned knowledge. Existing works largely fall into four paradigms, each limited when applied to disjoint physical sensors in PD assessment (Table~\ref{tab:related_comparison}). \textbf{Rehearsal-based methods} (e.g., MedCoSS \cite{ye2024continual}) reduce forgetting by replaying buffered legacy data, but this conflicts with patient privacy and hospital data-retention compliance. \textbf{Parameter Isolation techniques} (e.g., DRMN \cite{hegde2025modality}, CMR-MFN \cite{wang2023confusion}) allocate disjoint sub-networks or binary relevance maps for each modality. Although isolation protects old tasks, it blocks a unified cross-modal semantic boundary and therefore limits positive transfer. Finally, \textbf{Pre-trained Foundation Adaptation} \cite{liu2025continual, zhang2025contrastive} performs parameter-efficient fine-tuning over massive frozen backbones, which is impractical for wearable gait assessment because large paired pathological datasets for heterogeneous sensors are unavailable and heavy models conflict with edge-device constraints.

Latent projection and alignment are more suitable for our setting because they preserve a shared semantic manifold without full rehearsal or model isolation. Frameworks such as HarMI \cite{zhang2021harmi} and Harmony \cite{song2025harmony} are effective when modalities share homologous topologies (e.g., RGB, Depth, and IR images). However, when applied to disjoint clinical sensors (e.g., 1D temporal kinematics vs. 2D spatial kinetics), they expose two unresolved weaknesses.

First, for \textbf{modality-specific feature initialization}, current methods do not reliably anchor new sensor priors before alignment. HarMI \cite{zhang2021harmi} uses independent feature extractors for heterogeneous inputs, preserving modality separation but preventing a unified semantic boundary. Conversely, Harmony \cite{song2025harmony} relies on shallow, often linear, aggregators to map raw inputs into a shared backbone. However, a simple linear projections is structurally incapable of absorbing the severe incompatability between 2D walkway manifolds and 1D high-frequency IMU signals. Consequently, forcing immediate cross-modal KD from such uncalibrated mapping can therefore produce chaotic gradients that disrupt legacy decision boundaries.

Second, current architectures remain vulnerable to \textbf{cross-modal statistical interference} when bridging the \textit{modality gap} \cite{liang2022mind}. HarMI maximizes correlation through linear Canonical Correlation Analysis (CCA) \cite{zhang2021harmi}, which is insufficient for aligning highly non-linear clinical distributions. Harmony instead routes all streams through fully shared normalization layers, relying on architectural capacity to absorb statistical differences \cite{song2025harmony}. This may work for homologous visual modalities, over-parameterized networks, or concurrent multimodal/multi-task training \cite{zhou2026continual, zhou2026delve}, where all modalities are jointly observed and shared-statistics can be co-adapted. In compact MICL setting, however, modalities arrive sequentially without retaining historical raw data; under this constraint, forcing mismatched clinical modalities through shared normalization can overwrite historical statistical moments and trigger severe internal covariate shift \cite{ioffe2015batch}.

These two structural limitations, namely chaotic alignment gradients from weak initialization and catastrophic statistical coupling, highlight the necessity for a statistically decoupled yet parametrically shared architecture. This fundamental gap motivates our subsequent formal problem definition and the design of the MOSAIC framwork.

\begin{figure*}[!t]
\centering
\includegraphics[width=0.95\textwidth,trim=30 80 50 95,clip]{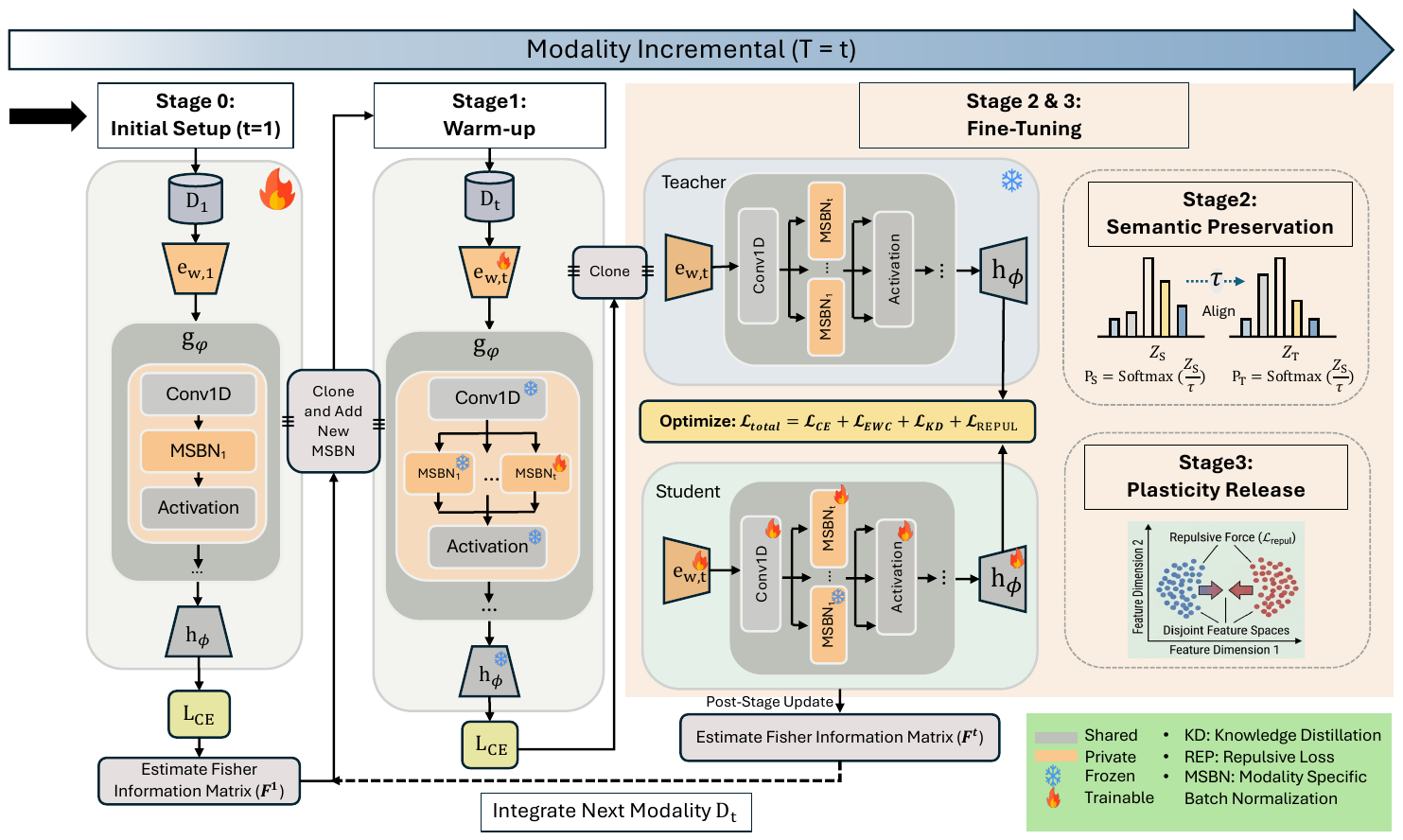}
\caption{Overview of the proposed modality-incremental continual learning pipeline for Parkinson's disease gait assessment. New sensing modalities are integrated sequentially without revisiting previous raw data. The framework combines modality-specific batch normalization to isolate variance-sensitive statistics, EWC-style regularization and distillation to preserve previously learned semantics, and a repulsive objective with curriculum scheduling to maintain plasticity under severe modality shift.}
\label{fig:mosaic_pipeline}
\end{figure*}

\section{Principle Design}
\label{sec:principle_design}

\subsection{Problem Setup: Continual Multimodal Gait Assessment}
\label{sec:problem_setup}
We model clinical PD gait assessment as a continual learning sequence over heterogeneous sensory domains. Let $\mathcal{T} = \{\mathcal{T}_1, \mathcal{T}_2, \dots, \mathcal{T}_N\}$ denote a sequence of $N$ diagnostic tasks. Unlike standard class-incremental learning, our setting follows a modality-incremental paradigm in which the diagnostic label space $\mathcal{Y}$ (e.g., healthy control vs. PD) remains fixed, while the input modality distribution changes across tasks.
Each task $\mathcal{T}_t$ introduces a dataset $\mathcal{D}_t = \{(\mathbf{x}_{i}^t, y_{i}^t)\}_{i=1}^{N_t}$, where $\mathbf{x}_{i}^t \in \mathcal{X}_t$ represents a time-series gait sequence collected from a distinct sensing modality $\mathcal{M}_t$, and $y_{i}^t \in \mathcal{Y}$ is the corresponding clinical label. The objective is to train a unified parameterized model $f_{\theta}: \mathcal{X} \rightarrow \mathcal{Y}$ sequentially from $\mathcal{T}_1$ to $\mathcal{T}_N$. During the training of task $\mathcal{T}_t$, the model only has access to the current dataset $\mathcal{D}_t$. 

Standard optimization of the empirical risk
\begin{equation}
\arg\min_{\theta} \; \mathbb{E}_{(\mathbf{x},y) \sim \mathcal{D}_t}\left[\mathcal{L}_{\mathrm{CE}}(f_{\theta}(\mathbf{x}), y)\right]
\end{equation}
inevitably leads to catastrophic forgetting of previous modalities $\mathcal{M}_{<t}$, where $\mathcal{L}_{\mathrm{CE}}$ is the cross-entropy loss. In multimodal PD assessment, this forgetting is fundamentally exacerbated by the extreme physical discrepancies between sensing domains (e.g., 1D temporal kinematics from IMUs vs. 2D spatial kinetics from clinical walkways). Thus, the framework must preserve historical decision boundaries and isolate heterogeneous feature statistics without storing raw legacy data.


\subsection{Overall Architecture: Statistics-Decoupled MSBN}
\label{sec:overall_architecture}
\begin{figure}[!t]
\centering
\includegraphics[width=0.98\linewidth]{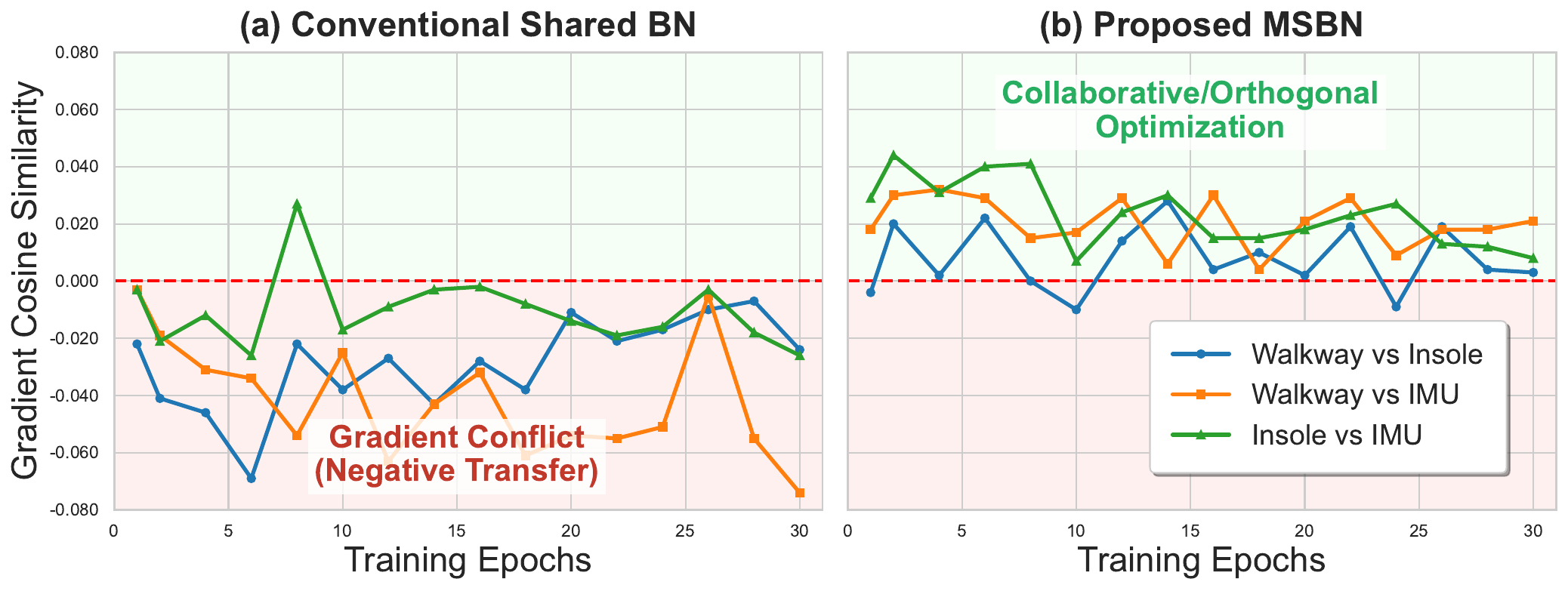}
\caption{Motivation analysis of shared BN under balanced joint training. Heterogeneous gait modalities are concatenated into shared BN, exposing mixed-statistics contamination and negative gradient interference in the shared convolutional backbone.}
\label{fig:motivation_msbn}
\vspace{-0.8em}
\end{figure}

As illustrated in Fig.~\ref{fig:mosaic_pipeline}, standard projection-based architectures force heterogeneous sensors through shared normalization layers. This induces cross-modal gradient interference, as incompatible statistical moments ($\mu, \sigma^2$) collide during backpropagation. As empirically demonstrated in Fig.~\ref{fig:motivation_msbn}(a), under Shared BN, the cosine similarity of gradients between disjoint modalities persistently drops below zero, providing definitive evidence of negative transfer. To mitigate this, we propose a statistics-decoupled MSBN architecture that explicitly isolates modality-specific statistics while preserving a shared semantic manifold. The unified network is composed of three core components:
\begin{equation}
f_{\theta} = h_{\phi} \circ g_{\psi} \circ e_{\omega_t}
\end{equation}
where $e_{\omega_t}$ denotes the modality-specific encoder for task $t$. $g_{\psi}$ represents a universal feature extractor for shared semantic modeling. Crucially, to insulate the network from cross-modal contamination, $g_{\psi}$ employs Modality-Specific Batch Normalization (MSBN). MSBN allocates independent affine parameters $(\gamma_t, \beta_t)$ and running statistics for each modality, preserving a common set of convolutional filters ($\psi$) without forcing them to process corrupted, mixed-variance gradients. As shown in Fig.~\ref{fig:motivation_msbn}(b), this structural decoupling successfully prevents gradient conflict, restoring orthogonal and collaborative optimization trajectories. Finally, $h_{\phi}$ is the shared classification head that aligns the abstract semantic features across modalities.

\subsection{Multi-Stage Optimization Strategy}
\label{sec:optimization_strategy}
Updating the shared backbone $g_{\psi}$ and classification head $h_{\phi}$ while integrating a novel sensor modality exposes the network to high-entropy gradient noise. To balance Modality-Specific Warm-Up, Semantic Preservation, and Plasticity Recovery, we adopt a three-stage optimization trajectory (summarized in Algorithm~\ref{alg:micl_training}).

\subsubsection{Stage 1: Modality-Specific Warm-Up}
When transitioning between physically disjoint domains, the newly initialized encoder $e_{\omega_t}$ and its MSBN parameters $(\gamma_t, \beta_t)$ lack valid mappings for the novel input scale. If the entire network is updated immediately, the uncalibrated encoder acts as a "Toxic Teacher" (Fig.~\ref{fig:toxic_teacher}), generating chaotic representations that permanently damage legacy weights.

Before cross-modal consolidation begins, the network must calibrate its sensor-specific statistics. Stage~1 performs Modality-Specific Warm-Up using a targeted cross-entropy objective. The shared parameters $\{ \psi, \phi \}$ are strictly frozen, restricting optimization solely to the novel encoder and its designated MSBN parameters:
\begin{equation}
\begin{aligned}
\arg\min_{\omega_t, \gamma_t, \beta_t} & \; \mathbb{E}_{(\mathbf{x}^t,y^t) \sim \mathcal{D}_t} \Big[ \\
& \mathcal{L}_{\mathrm{CE}}^{\mathrm{warmup}}\Big(h_{\phi}\big(g_{\psi}(e_{\omega_t}(\mathbf{x}^t), \gamma_t, \beta_t)\big), y^t\Big) \Big]
\label{eq:phase1_loss}
\end{aligned}
\end{equation}
By independently shifting and scaling the incoming feature distribution, this stage bounds the output norm and drops the pseudo-teacher entropy ($\mathcal{H}(p_t)$) and CE, anchoring a stable, legitimate statistical baseline without altering shared semantic weights (Fig.~\ref{fig:toxic_teacher}).

\subsubsection{Stage 2: Semantic Preservation}
Once Stage~1 converges, the anchored Modality-Specific Warm-Up model is cloned to form the frozen teacher network, $f_{\text{teacher}}$. The shared components $\{\psi, \phi\}$ are unfrozen, and Stage~2 performs Semantic Preservation by combining parameter regularization with logit distillation \cite{kirkpatrick2017overcoming,li2017learning}:
\begin{equation}
\mathcal{L}_{\mathrm{KD}} = \mathcal{D}_{\mathrm{KL}} \left( \sigma\left(\frac{z_{\text{s}}}{T}\right) \parallel \sigma\left(\frac{z_{\text{t}}}{T}\right) \right)
\label{eq:kd_loss}
\end{equation}
where $\sigma$ is the softmax function, $z_{\text{s}}$ and $z_{\text{t}}$ denote the student and teacher output logits, respectively, and $T$ is the temperature scalar. $\mathcal{L}_{\mathrm{KD}}$ transfers structural knowledge using only current-task data, while Elastic Weight Consolidation ($\mathcal{L}_{\mathrm{EWC}}$) penalizes deviations in parameters deemed critical to earlier modalities.

\subsubsection{Stage 3: Plasticity Recovery}
While $\mathcal{L}_{\mathrm{KD}}$ effectively preserves legacy structures, standard KD implicitly imposes rigid geometric constraints on the shared latent space. Because logits are linear projections of latent features, forcing exact probability matching ($z_{\text{s}} \approx z_{\text{t}}$) pressures the student's embedding to unilaterally replicate the exact spatial coordinates of the frozen teacher. For fundamentally incompatible clinical modalities, this strict spatial overlap restricts the network's capacity to model the unique topological structure of the incoming sensor, stifling plasticity.

To accommodate structural heterogeneity during Plasticity Recovery, we introduce a repulsive loss that actively maintains necessary topological boundaries. Rather than enforcing absolute spatial collapse, it dictates a controlled geometric margin between the student ($f_{\text{s}}$) and teacher ($f_{\text{t}}$) latent representations:
\begin{equation}
\mathcal{L}_{\mathrm{REP}} = \max\Big(0, \frac{f_{\text{s}} \cdot f_{\text{t}}}{\|f_{\text{s}}\|_2 \|f_{\text{t}}\|_2} - m_{\mathcal{T}}\Big)
\label{eq:repulsive_loss}
\end{equation}
where $m_{\mathcal{T}} \in [0,1]$ is a static \textit{Topological Margin}. By penalizing representations only when they become excessively entangled (i.e., cosine similarity $> m_{\mathcal{T}}$), the repulsive loss prevents destructive collapse while allowing independent latent capacity to emerge for modality-specific physics.

\subsubsection{Curriculum Scheduling for Plasticity Recovery}
To mitigate gradient conflict between semantic consolidation (KD) and spatial expansion (Repulsion), we introduce a continuous asymmetric curriculum. The objective weights are dynamically scheduled as:
\begin{equation}
\begin{aligned}
\lambda_{\mathrm{REP}}(t) &= \lambda_{\mathrm{REP}}^{\max} \cdot t^\gamma, \\
\lambda_{\mathrm{KD}}(t) &= \lambda_{\mathrm{KD}}^{\min} + \Delta\lambda_{\mathrm{KD}} \cdot (1 - t^\gamma),
\end{aligned}
\label{eq:curriculum_schedule}
\end{equation}
where $t \in [0,1]$ denotes the normalized training progress within the current task, $\Delta\lambda_{\mathrm{KD}} = \lambda_{\mathrm{KD}}^{\max} - \lambda_{\mathrm{KD}}^{\min}$, and $\gamma > 1$ is the curriculum exponent controlling the transition convexity.

This enforces a \textit{stabilize-then-expand} optimization trajectory. In early epochs, the distillation term dominates, encouraging stable Semantic Preservation. As training proceeds, the repulsive term progressively increases, allowing the model to allocate additional capacity for the incoming modality after the shared representations have stabilized. 

The final composite objective for Stages 2 and 3 is:
\begin{equation}
\mathcal{L}_{\mathrm{total}} = \mathcal{L}_{\mathrm{CE}}^{\mathrm{current}} + \mathcal{L}_{\mathrm{EWC}} + \lambda_{\mathrm{KD}}(t)\mathcal{L}_{\mathrm{KD}} + \lambda_{\mathrm{REP}}(t) \mathcal{L}_{\mathrm{REP}}
\label{eq:total_loss}
\end{equation}

\begin{algorithm}[t]
\caption{Training of MOSAIC}
\label{alg:micl_training}
\begin{algorithmic}[1]
\REQUIRE Sequential datasets $\{\mathcal{D}_t\}_{t=1}^{N}$, modality encoders $\{e_{\omega_t}\}$, shared backbone $g_{\psi}$ with statistics-decoupled MSBN, classifier $h_{\phi}$
\FOR{$t = 1$ to $N$}
    \IF{$t = 1$}
        \STATE Train on $\mathcal{D}_1$ with $\mathcal{L}_{\mathrm{CE}}$ and estimate Fisher statistics
    \ENDIF
    \STATE Attach $e_{\omega_t}$ and initialize MSBN parameters $(\gamma_t, \beta_t)$
    \STATE \textbf{Stage 1: Modality-Specific Warm-Up}
    \STATE Freeze $g_{\psi}$ and $h_{\phi}$; update only $\omega_t$, $\gamma_t$, and $\beta_t$ with $\mathcal{L}_{\mathrm{CE}}^{\mathrm{warmup}}$
    \STATE \textbf{Stage 2: Semantic Preservation}
    \STATE Clone Modality-Specific Warm-Up model as $f_{\text{teacher}}$ and unfreeze $g_{\psi}, h_{\phi}$
    \FOR{consolidation epochs $k$}
        \STATE Compute $\lambda_{\mathrm{KD}}(k)$ and optimize $\mathcal{L}_{\mathrm{CE}}^{\mathrm{current}} + \mathcal{L}_{\mathrm{EWC}} + \lambda_{\mathrm{KD}}(k)\mathcal{L}_{\mathrm{KD}}$
    \ENDFOR
    \STATE \textbf{Stage 3: Plasticity Recovery}
    \FOR{late epochs $k$}
        \STATE Compute $\lambda_{\mathrm{KD}}(k)$ and $\lambda_{\mathrm{REP}}(k)$
        \STATE Optimize $\mathcal{L}_{\mathrm{CE}}^{\mathrm{current}} + \mathcal{L}_{\mathrm{EWC}} + \lambda_{\mathrm{KD}}(k)\mathcal{L}_{\mathrm{KD}} + \lambda_{\mathrm{REP}}(k)\mathcal{L}_{\mathrm{REP}}$
    \ENDFOR
    \STATE Update Fisher statistics and retain model for task $t{+}1$
\ENDFOR
\RETURN Final continual model $f_{\theta}$
\end{algorithmic}
\end{algorithm}

\begin{table*}[!t]
\centering
\caption{Main Results on Modality-Incremental Learning Scenarios. We compare our framework against standard Continual Learning baselines across three multimodal datasets and different modality arrival orders. ACC denotes average accuracy across all seen modalities after training is complete. BWT denotes Backward Transfer (negative values indicate forgetting). NAA denotes Normalized Average Accuracy.}
\label{tab:main_results}
\begin{adjustbox}{width=\textwidth}
\begin{tabular}{l|ccc|ccc|ccc}
\toprule
\multirow{4}{*}{\textbf{Method}} & \multicolumn{3}{c|}{\textbf{WearGait}} & \multicolumn{3}{c|}{\textbf{FBG}} & \multicolumn{3}{c}{\textbf{FOG}} \\
& \multicolumn{3}{c|}{(Walkway / Insole / IMU)} & \multicolumn{3}{c|}{(Skel / Angular / GRF)} & \multicolumn{3}{c}{(Skel / Gyro / Acc)} \\
\cmidrule(lr){2-4} \cmidrule(lr){5-7} \cmidrule(lr){8-10}
& \textbf{F1 (\%) $\uparrow$} & \textbf{BWT (\%) $\uparrow$} & \textbf{NAA (\%) $\uparrow$} & \textbf{F1 (\%) $\uparrow$} & \textbf{BWT (\%) $\uparrow$} & \textbf{NAA (\%) $\uparrow$} & \textbf{F1 (\%) $\uparrow$} & \textbf{BWT (\%) $\uparrow$} & \textbf{NAA (\%) $\uparrow$} \\
\midrule
\multicolumn{10}{l}{\textit{\textbf{Order-Independent Oracle Bounds}}} \\
Oracle M1 & 80.20 & - & - & 67.29 & - & - & 36.92 & - & -\\
Oracle M2 & 62.40 & - & - & 62.78 & - & - & 42.67 & - & -\\
Oracle M3 & 78.27 & - & - & 56.92 & - & - & 42.47 & - & -\\
TRIP \cite{zeng2025towards} & 83.42 & - & - & 70.26 & - & - & 45.52 & - & - \\
\midrule
\textit{\textbf{Order A}} & \multicolumn{3}{c|}{Walkway $\to$ Insole $\to$ IMU} & \multicolumn{3}{c|}{Skel $\to$ Angular $\to$ GRF} & \multicolumn{3}{c}{Skel $\to$ Gyro $\to$ Acc} \\
\cmidrule(lr){2-4} \cmidrule(lr){5-7} \cmidrule(lr){8-10}
Fine-Tuning & 57.15 & -23.87 & 79.25 & 54.53 & -7.87 & 87.97 & 31.77 & -11.60 & 77.30 \\
EWC \cite{kirkpatrick2017overcoming} & 64.81 & -11.54 & 92.73 & 53.87 & -7.97 & 86.90 & 30.62 & -16.77 & 74.88 \\
LwF \cite{li2017learning} & \second{68.17} & -6.29 & \second{93.10} & 53.30 & -4.34 & 85.77 & 31.03 & -15.53 & 75.12 \\
LwI \cite{chenlearning} & 34.14 & -21.14 & 46.96 & 44.03 & -7.95 & 70.88 & 34.45 & -2.77 & 85.39 \\
MedCoSS \cite{ye2024continual} & 45.45 & \second{-4.92} & 61.89 & \second{56.17} & \second{-1.38} & \second{90.30} & 35.50 & \best{-0.46} & 83.84 \\
Harmony \cite{song2025harmony} & 58.63 & -14.76 & 80.21 & 46.52 & -11.78 & 75.19 & \second{36.59} & -13.29 & \second{89.70} \\
DRMN \cite{hegde2025modality} & 52.05 & -22.49 & 72.82 & 43.57 & -8.65 & 71.73 & 31.76 & -14.88 & 77.73 \\
\textbf{Ours} & \best{73.22} & \best{-1.34} & \best{99.76} & \best{60.29} & \best{0.64} & \best{96.79} & \best{41.14} & \second{-2.74} & \best{100.82} \\
\midrule
\textit{\textbf{Order B}} & \multicolumn{3}{c|}{Walkway $\to$ IMU $\to$ Insole} & \multicolumn{3}{c|}{GRF $\to$ Angular $\to$ Skel} & \multicolumn{3}{c}{Acc $\to$ Gyro $\to$ Skel} \\
\cmidrule(lr){2-4} \cmidrule(lr){5-7} \cmidrule(lr){8-10}
Fine-Tuning & 61.56 & -17.06 & 84.82 & \second{56.30} & \second{-5.44} & \second{90.13} & 31.35 & -13.35 & 78.04 \\
EWC \cite{kirkpatrick2017overcoming} & 66.33 & -9.28 & 90.82 & 52.34 & -8.90 & 83.67 & 31.15 & -13.42 & 77.76 \\
LwF \cite{li2017learning} & \second{69.76} & -6.01 & \second{95.35} & 52.38 & -8.87 & 83.74 & \second{36.59} & -6.95 & \second{90.62} \\
LwI \cite{chenlearning} & 33.52 & -22.76 & 45.98 & 38.49 & -10.36 & 61.88 &  34.79 & -4.75 & 86.06 \\
MedCoSS \cite{ye2024continual} & 52.41 & \second{-2.78} & 72.41 & 53.76 & -6.30 & 85.92 & 32.35 & \best{-2.86} & 79.67 \\
Harmony \cite{song2025harmony} & 46.85 & -25.20 & 65.11 & 45.28 & -13.31 & 72.31 & 35.25 & -10.76 & 87.27  \\
DRMN \cite{hegde2025modality} & 50.37 & -24.47 & 70.83 & 44.75 & -12.26 & 70.90 & 31.94 & -15.05 & 80.11  \\
\textbf{Ours} & \best{73.04} & \best{-1.40} & \best{99.53} & \best{58.42} & \best{0.11} & \best{93.99} & \best{40.24} & \second{-3.95} & \best{99.46} \\
\bottomrule
\end{tabular}
\end{adjustbox}
\end{table*}

\section{Implementation}
\subsection{Training and Validation Details}
All experiments are implemented in PyTorch and rigorously evaluated using 5-fold cross-validation. Macro-F1 serves as the primary validation metric for model selection and early stopping. Models are optimized using AdamW, with base learning rates ($10^{-4}$ to $10^{-3}$) and batch sizes ($32$ to $128$) tailored to the specific scale and sensor physics of each dataset. Training strictly follows our proposed three-stage procedure: Modality-Specific Warm-Up, Semantic Preservation, and Plasticity Recovery within the statistics-decoupled MSBN architecture. Comprehensive implementation details, including statistics-decoupled MSBN mechanisms and explicit continual learning constraint configurations, are provided in the Supplementary Appendix, Sec.~III.

\subsection{Baselines}
To evaluate MOSAIC across practical continual-learning paradigms and oracle bounds, we benchmark against three groups of baselines. First, \textit{Naive Fine-Tuning} serves as the lower bound by sequentially updating the model without anti-forgetting constraints, while \textit{Specialist Training} and \textit{Joint Training (TRIP \cite{zeng2025towards})} provide single-modality and concurrent-multimodal upper bounds. Second, \textit{EWC} \cite{kirkpatrick2017overcoming} and \textit{LwF} \cite{li2017learning} represent standard parameter-regularization and logit-distillation architectures. Third, we include SOTA MICL frameworks: \textit{LwI} \cite{chenlearning} and \textit{DRMN} \cite{hegde2025modality} for isolation-based learning, \textit{Harmony} \cite{song2025harmony} for latent-alignment modality-incremental learning, and \textit{MedCoSS} \cite{ye2024continual} representing rehearsal-based memory models.

\section{Evaluation and Experiments}
\subsection{Evaluation Metrics}
We use standard metrics commonly adopted in continual learning and medical classification. Macro-F1 is reported as the primary classification metric because it is less biased by class imbalance than accuracy. Let $T$ denote the number of modality tasks, and let $R_{i,j}$ be the Macro-F1 on task $j$ after training has completed on task $i$. To quantify forgetting, we report Backward Transfer (BWT) \cite{lopez2017gradient}:
\begin{equation}
BWT = \frac{1}{T-1} \sum_{j=1}^{T-1} (R_{T,j} - R_{j,j}).
\end{equation}
A less negative BWT indicates better retention of previous modalities. However, BWT alone is insufficient in our setting because different sensor modalities can have substantially different specialist performance. A method may show low forgetting while still performing far below the attainable performance of a difficult modality, or may appear worse simply because one modality has a lower absolute performance ceiling. 

Therefore, we also use Normalized Average Accuracy (NAA) \cite{kemker2018measuring}, which normalizes final continual performance by the corresponding single-modality specialist. Let $b_j$ denote the specialist performance for task $j$:
\begin{equation}
\Omega = \frac{1}{T} \sum_{j=1}^{T} \frac{R_{T,j}}{b_j}.
\end{equation}
Higher NAA indicates that the continual model better approaches the modality-specific upper bounds, making it complementary to BWT for evaluating both retention and final normalized performance.

\subsection{PD Gait Multimodal Datasets}
We conducted evaluations on three publicly available multimodal PD gait datasets. Following the modality-incremental setting, each dataset is organized into sequential modality phases while maintaining a unified clinical label space. Across datasets, the evaluation covers three representative gait sensing modalities: pose/kinematic trajectories, force or pressure measurements, and wearable inertial signals.

\noindent\textbf{WearGait Dataset} \cite{kontson2024wearables}: This dataset provides synchronized walkway, sensorized insole, and IMU data from 98 people with PD and 83 age-matched controls. We structure it into three modality phases: pressure-walkway signals, plantar-pressure insole signals, and wearable IMU signals. All streams are resampled to 30 Hz, aligned, imputed for missing values, and segmented into fixed-length windows for training and evaluation. Each phase performs PD vs. control gait classification.

\noindent\textbf{FBG Dataset} \cite{shida2023public}: This dataset includes synchronized full-body motion capture and force-plate recordings from 26 PD subjects walking overground in ON/OFF medication states. We construct three modality tasks from synchronized motion-capture and force-plate recordings: skeleton trajectories (Skel), joint-angular features (Angular), and ground-reaction force signals (GRF).

\noindent\textbf{FOG Dataset} \cite{ribeiro2022public}: This dataset contains video and IMU recordings from 35 PD subjects performing turning-in-place, with clinical freezing-of-gait annotations. We use video-derived lower-limb skeletons and IMU streams (gyroscope (Gyro) and accelerometer signals (Acc)) to form three modality tasks. Skeleton sequences are normalized and fixed to $F=101$ frames, while IMU trials are temporally segmented and aligned with the corresponding skeleton segments.

\subsection{Main Results}
Table~\ref{tab:main_results} summarizes the comparative performance under MICL setting across two arrival sequences (Orders A and B) per dataset to account for task-order sensitivity \cite{lai2025order}. MOSAIC consistently establishes SOTA performance, achieving peak F1 scores across datasets. Concurrently, MOSAIC yields near-ideal NAA ($\ge$ 96.79\% across all evaluations), indicating that the continual network closely approximates the performance ceilings of independent single-modality specialists.

The performance degradation of existing baselines empirically validates our architectural critiques. Frameworks relying on shared normalization spaces, such as Harmony, suffer severe internal covariate shift and feature corruption; for example, on WearGait (Order A), Harmony achieves an F1 of only 58.63\% accompanied by massive forgetting (BWT $-14.76\%$). Conversely, parameter-isolation methods like DRMN attempt to bypass interference through structural decoupling, but this completely blocks positive semantic transfer, resulting in severely stunted final capacity (F1 52.05\%, BWT $-22.49\%$). Standard distillation baselines (e.g., LwF) mitigate some forgetting (BWT $-6.29\%$) but fail to calibrate statistical shifts prior to alignment. By shielding historical topologies via the statistics-decoupled MSBN architecture and safely expanding the representation space through Plasticity Recovery, MOSAIC successfully breaks the plasticity-retention trade-off, minimizing forgetting while maximizing representational capacity.

\begin{table}[!t]
\caption{Ablation analysis of MOSAIC on WearGait. The component columns follow the proposed design: Stage~1 performs Modality-Specific Warm-Up for the new encoder and its MSBN statistics within the statistics-decoupled MSBN architecture, Stage~2 performs Semantic Preservation with EWC+KD, and Stage~3 performs Plasticity Recovery through curriculum-guided repulsion. Bold and underlining indicate the best and second-best results, respectively.}
\label{tab:ablation_components}
\resizebox{\columnwidth}{!}{%
\begin{tabular}{lccccc}
\toprule
\multirow{1}{*}{\textbf{Model Variant}} & \textbf{Stage 1} & \textbf{Stage 2} & \textbf{Stage 3} & \multirow{1}{*}{\textbf{BWT (\%) $\uparrow$}} & \multirow{1}{*}{\textbf{NAA (\%) $\uparrow$}} \\
\midrule
Naive Fine-Tuning &  &  &  & -22.13 & 81.33 \\
+ EWC+KD (Standard CL) &  & $\checkmark$ &  & -4.73 & 94.42 \\
+ Modality-Specific Warm-Up & $\checkmark$ & $\checkmark$ &  & \second{-2.81} & \second{98.71} \\
+ Curriculum REP. (MOSAIC) & $\checkmark$ & $\checkmark$ & $\checkmark$ & \textbf{-1.73} & \textbf{99.79} \\
\bottomrule
\end{tabular}%
}
\end{table}

\subsection{Ablation Study}
Table~\ref{tab:ablation_components} validates the contribution of each component on WearGait using the same naming as the Principle Design section. Naive fine-tuning suffers severe forgetting (BWT $-22.13\%$, NAA $81.33\%$). Adding Stage~2 Semantic Preservation with EWC+KD reduces forgetting to $-4.73\%$ and raises NAA to $94.42\%$, confirming the importance of retaining legacy decision boundaries. Incorporating Stage~1 Modality-Specific Warm-Up further improves BWT/NAA to $-2.81\%$/$98.71\%$, showing that the new encoder and MSBN statistics must be calibrated before consolidation. Finally, the full MOSAIC configuration with Stage~3 Plasticity Recovery achieves the best performance (BWT $-1.73\%$, NAA $99.79\%$), indicating that repulsion is most effective when introduced through the stabilize-then-expand curriculum. Step-wise ablation results and additional MSBN-augmented EWC/LwF comparisons are provided in Supplementary Appendix, Secs.~III and~II, respectively.

\begin{figure}[!t]
\centering
\resizebox{0.82\columnwidth}{!}{%
\begin{tikzpicture}
    \begin{axis}[
        width=0.74\columnwidth,
        height=0.48\columnwidth,
        scale only axis,
        xmin=-0.25, xmax=0.45,
        ymin=-2.5, ymax=-0.5,
        axis y line*=left,
        xlabel={Repulsive Margin ($m$)},
        ylabel={BWT (\%)},
        ylabel style={text=academicblue, font=\footnotesize\bfseries},
        xlabel style={font=\footnotesize\bfseries},
        yticklabel style={text=academicblue, font=\footnotesize},
        grid=major,
        grid style={dashed, gray!40},
        xticklabel style={font=\footnotesize},
        legend style={at={(0.03,0.8)}, anchor=south west, fill=white, fill opacity=0.9, draw opacity=1, text opacity=1, font=\scriptsize},
    ]
        \fill[yellow!20, opacity=0.6] (axis cs:0.25,-2.5) rectangle (axis cs:0.35,-0.5);
        \draw[gray, thick, dash dot] (axis cs:0.3,-2.5) -- (axis cs:0.3,-0.5)
            node[pos=0.95, right, font=\footnotesize\bfseries, text=black] {Optimal};

        \addplot[color=academicblue, mark=*, mark size=2.2pt, line width=1.5pt] coordinates {
            (-0.2, -2.01)
            (0.0, -2.19)
            (0.1, -2.33)
            (0.2, -1.14)
            (0.3, -1.73)
            (0.4, -1.28)
        };
        \addlegendentry{BWT $\uparrow$}
    \end{axis}

    \begin{axis}[
        width=0.74\columnwidth,
        height=0.48\columnwidth,
        scale only axis,
        xmin=-0.25, xmax=0.45,
        ymin=0.980, ymax=1.000,
        axis y line*=right,
        axis x line=none,
        ylabel={NAA},
        ylabel style={text=academicred, font=\footnotesize\bfseries},
        ytick={0.980, 0.985, 0.990, 0.995, 1.000},
        yticklabel style={text=academicred, font=\footnotesize},
        yticklabel={\pgfmathprintnumber[fixed, precision=3]{\tick}},
        legend style={at={(0.03,0.66)}, anchor=south west, fill=white, fill opacity=0.9, draw opacity=1, text opacity=1, font=\scriptsize},
    ]
        \addplot[color=academicred, mark=triangle*, mark size=2.6pt, line width=1.5pt, dashed] coordinates {
            (-0.2, 0.9871)
            (0.0, 0.9862)
            (0.1, 0.9867)
            (0.2, 0.9904)
            (0.3, 0.9979)
            (0.4, 0.9923)
        };
        \addlegendentry{NAA $\uparrow$}
    \end{axis}
\end{tikzpicture}%
}
\caption{Sensitivity analysis of the repulsive margin $m$. The selected setting ($m=0.3$) yields the highest normalized average accuracy while maintaining favorable backward transfer, indicating a robust trade-off between modality separation and knowledge retention.}
\label{fig:ablation_margin}
\end{figure}
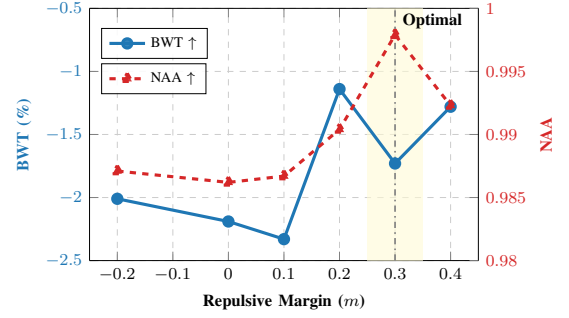
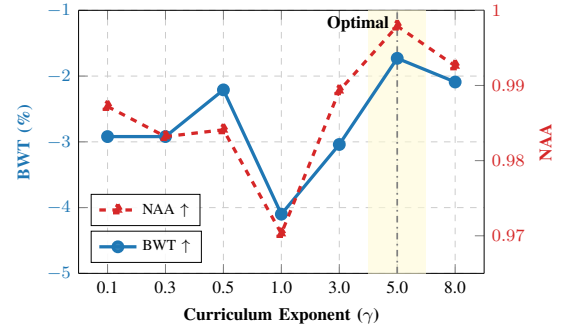
\begin{figure}[!t]
\centering
\resizebox{0.82\columnwidth}{!}{%
\begin{tikzpicture}
    \begin{axis}[
        width=0.74\columnwidth,
        height=0.48\columnwidth,
        scale only axis,
        xmin=0.5, xmax=7.5,
        ymin=-5.0, ymax=-1.0,
        axis y line*=left,
        xlabel={Curriculum Exponent ($\gamma$)},
        ylabel={BWT (\%)},
        ylabel style={text=academicblue, font=\footnotesize\bfseries},
        xlabel style={font=\footnotesize\bfseries},
        yticklabel style={text=academicblue, font=\footnotesize},
        xtick={1, 2, 3, 4, 5, 6, 7},
        xticklabels={0.1, 0.3, 0.5, 1.0, 3.0, 5.0, 8.0},
        grid=major,
        grid style={dashed, gray!40},
        xticklabel style={font=\footnotesize},
        legend style={at={(0.03,0.04)}, anchor=south west, fill=white, fill opacity=0.9, draw opacity=1, text opacity=1, font=\scriptsize},
    ]
        \fill[yellow!20, opacity=0.6] (axis cs:5.5,-5.0) rectangle (axis cs:6.5,-1.0);
        \draw[gray, thick, dash dot] (axis cs:6,-5.0) -- (axis cs:6,-1.0)
            node[pos=0.95, left, font=\footnotesize\bfseries, text=black] {Optimal};

        \addplot[color=academicblue, mark=*, mark size=2.2pt, line width=1.5pt] coordinates {
            (1, -2.92)
            (2, -2.92)
            (3, -2.21)
            (4, -4.10)
            (5, -3.04)
            (6, -1.73)
            (7, -2.09)
        };
        \addlegendentry{BWT $\uparrow$}
    \end{axis}

    \begin{axis}[
        width=0.74\columnwidth,
        height=0.48\columnwidth,
        scale only axis,
        xmin=0.5, xmax=7.5,
        ymin=0.965, ymax=1.000,
        axis y line*=right,
        axis x line=none,
        ylabel={NAA},
        ylabel style={text=academicred, font=\footnotesize\bfseries},
        yticklabel style={text=academicred, font=\footnotesize},
        legend style={at={(0.03,0.18)}, anchor=south west, fill=white, fill opacity=0.9, draw opacity=1, text opacity=1, font=\scriptsize},
    ]
        \addplot[color=academicred, mark=triangle*, mark size=2.6pt, line width=1.5pt, dashed] coordinates {
            (1, 0.9872)
            (2, 0.9832)
            (3, 0.9841)
            (4, 0.9703)
            (5, 0.9893)
            (6, 0.9979)
            (7, 0.9926)
        };
        \addlegendentry{NAA $\uparrow$}
    \end{axis}
\end{tikzpicture}%
}
\caption{Sensitivity analysis of the curriculum exponent $\gamma$. The selected setting ($\gamma=5.0$) provides the best balance between backward transfer and normalized average accuracy, supporting a gradual schedule for introducing repulsive separation after semantic stabilization.}
\label{fig:ablation_gamma}
\end{figure}

\subsection{Hyperparameters Analysis}
\subsubsection{\texorpdfstring{Topological Margin ($m_{\mathcal{T}}$)}{Topological Margin (mT)}}
\label{topological_margin}
Fig.~\ref{fig:ablation_margin} shows that the model is generally robust to the repulsive margin $m$, with NAA remaining above $0.986$ across tested settings. Very small margins ($m \leq 0.1$) increase forgetting, whereas $m=0.3$ achieves the best normalized performance (NAA $0.9979$) with competitive BWT ($-1.73\%$). We therefore use $m=0.3$ as the default trade-off between retention and final performance.

\subsubsection{\texorpdfstring{Curriculum Exponent ($\gamma$)}{Curriculum Exponent (gamma)}}
Fig.~\ref{fig:ablation_gamma} evaluates the curriculum exponent $\gamma$, which controls how gradually the repulsive objective is introduced. A linear schedule ($\gamma=1.0$) disrupts retention (BWT $-4.10\%$, NAA $0.9703$), while $\gamma=5.0$ provides the best overall balance, reaching BWT $-1.73\%$ and NAA $0.9979$. This supports a curriculum that first consolidates semantic knowledge and then progressively strengthens inter-modality separation.


\section{Discussion}
The results demonstrate that our design can reduce forgetting under modality-incremental gait assessment. Nevertheless, a limitation of current framework is that the repulsive objective uses a fixed topological margin after it is selected. A static margin provides a simple and interpretable separation criterion, but it may not fully reflect the varying degrees of compatibility among modality pairs, such as closely related pressure-based sensors versus more heterogeneous inertial and force-plate measurements. Future work should explore adaptive repulsion strategies in which the margin or repulsion strength is dynamically estimated from inter-modality similarity or validation feedback during continual adaptation.

\section{Conclusion}
In this work, we identified and resolved fundamental optimization bottlenecks in MICL for PD gait assessment. We demonstrated that existing alignment paradigms suffer from "Toxic Teacher" gradient corruption and statistical overwriting when applied to disjoint physical sensors. To address this, we proposed MOSAIC, a novel framework that coordinates a statistics-decoupled MSBN architecture with Modality-Specific Warm-Up, Semantic Preservation, and Plasticity Recovery. Extensive evaluations across three clinical datasets confirm that MOSAIC successfully shields historical knowledge from catastrophic forgetting while recovering necessary plasticity for novel modalities, establishing a robust foundation for evolving clinical AIoT diagnostic systems.

\makeatletter

\def\thebibliography#1{\section*{References}%
    \footnotesize \vskip 0.3\baselineskip plus 0.1\baselineskip minus 0.1\baselineskip%
    \list{\@biblabel{\@arabic\c@enumiv}}%
    {\settowidth\labelwidth{\@biblabel{#1}}%
    \leftmargin\labelwidth
    \advance\leftmargin\labelsep\relax
    \itemsep 0pt plus .5pt\relax%
    \usecounter{enumiv}%
    \let\p@enumiv\@empty
    \renewcommand\theenumiv{\@arabic\c@enumiv}}%
    \let\@IEEElatexbibitem\bibitem%
    \def\bibitem{\@IEEEbibitemprefix\@IEEElatexbibitem}%
    \def\newblock{\hskip .11em plus .33em minus .07em}%
    \if@technote\sloppy\clubpenalty4000\widowpenalty4000\interlinepenalty100%
    \else\sloppy\clubpenalty4000\widowpenalty4000\interlinepenalty500\fi%
    \sfcode`\.=1000\relax}
\makeatother
\bibliographystyle{IEEEtran}
\bibliography{reference}

\end{document}